# Wastewater Pipe Condition Rating Model Using $K$-Nearest Neighbors

*Sai Nethra Betgeri, Louisiana Tech University, Ruston, LA*

*Shashank Reddy Vadyala, Louisiana Tech University, Ruston, LA*

*Dr. John C. Matthews, Director, TTC, Louisiana Tech University, Ruston, LA*

*Dr. Mahboubeh Madadi, Assistant Professor, San Jose State University, San Jose, CA*

*Dr. Greta Vladeanu, Data Scientist, Xylem, Columbia, MD*

**Abstract:**

Risk-based assessment in pipe condition mainly focuses on prioritizing the most critical assets by evaluating the risk of pipe failure. This paper's goal is to classify a comprehensive pipe rating model which is obtained based on a series of pipe physical, external, and hydraulic characteristics that are identified for the proposed methodology. The traditional manual method of assessing sewage structural conditions takes a long time. By building an automated process using $K$-Nearest Neighbors ($K$-NN), this study presents an effective technique to automate the identification of the pipe defect rating using the pipe repair data. First, we performed the Shapiro Wilks Test for 1240 data from the Dept. of Engineering & Environmental Services, Shreveport, Louisiana Phase 3 with 12 variables to determine if factors could be incorporated in the final rating. We then developed a $K$-Nearest Neighbors model to classify the final rating from the statistically significant factors identified in Shapiro Wilks Test. This classification process allows recognizing the worst condition of wastewater pipes that need to be replaced immediately. This comprehensive model is built according to the industry-accepted and used guidelines to estimate the overall condition. Finally, for validation purposes, the proposed model is applied to a small portion of a US wastewater collection system in Shreveport, Louisiana.

Keywords: Pipe rating, Shapiro Wilks Test, $K$-Nearest Neighbors (KNN), Failure, Risk analysis

# 1. Introduction

The underground pipeline system forms a significant part of the infrastructure in the United States because it includes thousands of miles of pipes. Sanitary sewage collects wastewater from public and private users as part of wastewater infrastructure systems(Ariaratnam et al., 2001). Approximately 800,000 miles of public sewage pipes and 500,000 miles of private sewer laterals are present(Lester & Farrar, 1979). By 2032, 56 million people are expected to use centralized treatment plants(Davies, Clarke, Whiter, & Cunningham, 2001; Davies, Clarke, Whiter, Cunningham, et al., 2001; O'REILLY et al., 1989). Water supply and sewer water pipeline is basic need for society, and their security and efficiency are of paramount importance to human health and economic development. However, a substantial portion of these vital systems are decades old and are plagued with deficiencies and inefficiencies, as shown in Figure 1. Using risk-based asset management, the most critical assets to take the most efficient course of action are identified by prioritizing the highest risk of failure by considering parameters such as pipe diameter, material, age, wall thickness(Aprajita, 2018). The four focus areas of a risk-based asset management program are (1) understanding the deterioration modes and mechanisms, (2) risk assessment and management, (3) condition assessment, and (4) asset renewal, i.e., repair/rehabilitation/replacement. Using the traditional method, the number of failures received by sewer management can increase rapidly, making pipe failure handling imperative. However, manually identifying and classifying those failures through pipe repair documents and extracting the information related to those pipe failures are challenging because they are voice-generated text or manually written by the inspectors by inspecting through closed-circuit television (CCTV) videos. This manual process can lead to many more committed errors (Sai Nethra Betgeri, 2021; Vadyala & Betgeri, 2021; Vadyala et al., 2021; G. Vladeanu & J. Matthews, 2019; G. J. Vladeanu & J. C. Matthews, 2019). CCTV inspection includes two effective methods: on-site video collection and off-site video assessment, as shown in Figure 2.

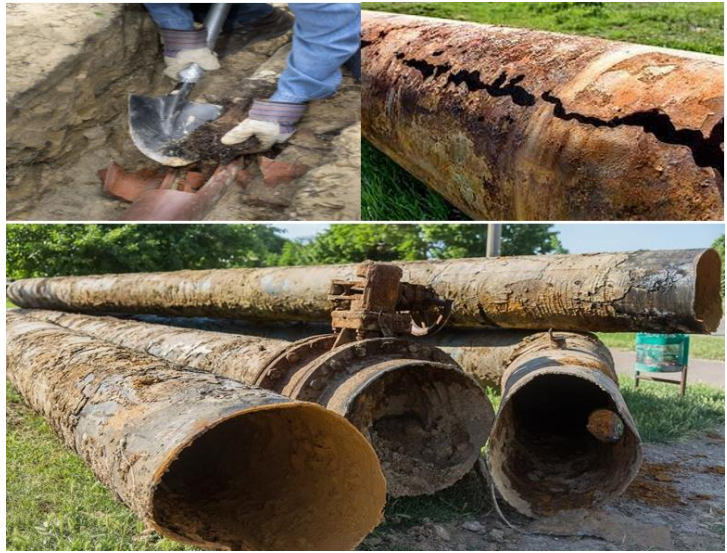

Figure 1: Different pipe deficiencies: leakage; partial blockage; deformation; corrosion, detachment.

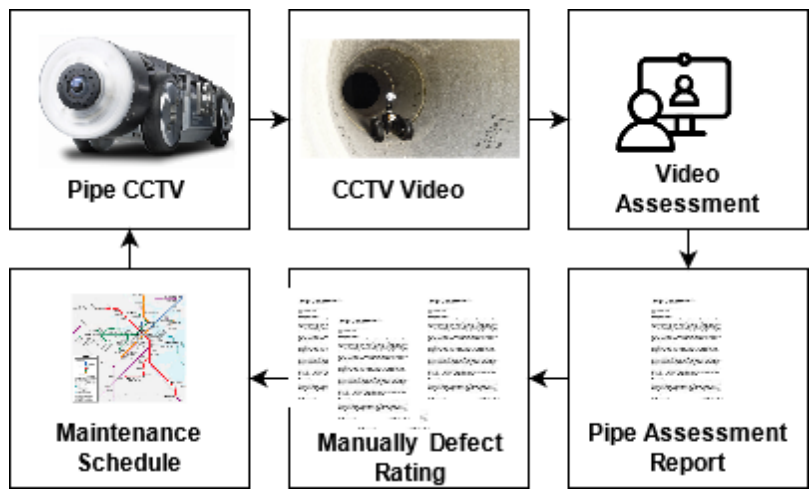

Figure 2: Overview of video assessment

Many ways for monitoring and reporting the state of sewage lines have been developed. Different approaches use different input parameters to compute a structural and/or operational condition grade. Creating a condition assessment system like this is to prioritize assets for future interventions like comprehensive inspections and renewal programs. A rating method typically indicates the pipe's condition on a 1–5 scale, with 1 being the best condition and 5 requiring immediate replacement(Angkasuwansiri & Sinha, 2015; Companies; Khazraeializadeh et al., 2014; McDonald & Zhao, 2020; Opila, 2011; Wirahadikusumah et al., 2001). The National Association of Sewer Service Companies (NASSCO) Pipeline Assessment Certification Program (PACP) technique, based on the Water Research Center (WRc) standards, is a standardized condition assessment approach that analyzes wastewater pipe conditions based on observable structural and operational faults in the United States(PACP, 2021; Rehab).

The PACP technique does not include any internal or exterior characteristics (such as pipe material, burial depth, soil characteristics, etc.) when determining the pipe's condition rating, rendering it inappropriate for pipe renewal decision-making(Opila, 2011; Opila & Attoh-Okine, 2011). As a result, in addition to the PACP defect ratings, other relevant factors, such as soil type, traffic loading, waste type, seismic zone related to external characteristics and structural score, operational & maintenance score, repair history related to hydraulic characteristics. Were assessed to provide a more precise assessment of the pipes' condition by the inspectors manually. PACP defect ratings are listed in Table 1, and comprehensive rating descriptions are listed in Table 2.

Table 1. PACP ratings and description

| PACP Ratings | Description |
| --- | --- |
| Defect rating 1 | Unlikely in the foreseeable future |
| Defect rating 2 | Rehabilitate or replace in 20 years or more |
| Defect rating 3 | Rehabilitate or replace in ten to twenty years. |
| Defect rating 4 | Rehabilitate or replace in five to ten years. |
| Defect rating 5 | Rehabilitate or replace in next five years |

Table 2. Comprehensive rating and description given by the inspector

| Comprehensive Ratings | Description |
| --- | --- |
| Defect rating 1 | Reassess in ten years |
| Defect rating 2 | Rehabilitate or replace in six to ten years |
| Defect rating 3 | Rehabilitate or replace in three to five years. |
| Defect rating 4 | Rehabilitate or replace in zero to two years. |
| Defect rating 5 | Rehabilitate or replace immediately. |

## 2. Factors Affecting Wastewater Pipe Condition

The degradation of wastewater pipes does not follow a predictable pattern and is influenced by various internal and external characteristics(Najafi & Kulandaivel, 2005). The most common causes that impact sewage pipe degradation may be classified into the following groups: construction variables, external parameters, and other factors. Sewer pipe diameter, pipe material, burial depth, pipe bedding, load transfer, pipe joint type and material, and sewer pipe connection are all construction variables(Davies, Clarke, Whiter, & Cunningham, 2001; Davies, Clarke, Whiter, Cunningham, et al., 2001). In addition, surface loading, ground conditions, groundwater level, and soil type are included in external parameters(Chughtai & Zayed, 2008; Yan & Vairavamoorthy, 2003). Finally, various factors include the type of waste carried, pipe age, sediment level, surcharge, and poor maintenance practices(Ennaouri & Fuamba, 2013).

## 3. Objective

This paper aims to automate the process of determining a comprehensive pipe rating through the development of an automated model that evaluates the overall condition of wastewater pipes segments. This is achieved by combining pipe characteristics, external characteristics, and hydraulic characteristics to be used by the utility department. The proposed model is designed to integrate previous research efforts and suggests an innovative method of using 12 factors encompassing physical characteristics, external conditions, and structural, operational, and hydraulic factors into assessing the overall condition of the pipe. We performed the Shapiro Wilks Test to perform hypothesis testing and determine the final factors that can be incorporated in the In Machine Learning, data satisfying Normal Distribution is beneficial for model building because many data in nature displays the bell-shaped curve when graphed. We finally developed a model with 10 factors identified from the Shapiro Wilks test for model building. The main goal of the developed model is to prioritize the worst condition assets for intervention planning in significantly less than 24 hours for classifying the rating through automation, where the inspector takes days or weeks to go through those pipes.

## 4. Methodology
### 4.1 Data Set:

A total of 3100 pipe data totaling approximately 285km (935,703 ft) is given. For this study, a total length of roughly 47 km (154,060 ft) of 200 mm (8 in.) diameter vitrified clay (VC) pipe, totaling 1240 pipe segments, were selected. Information such as diameter, depth, length of the pipes is given in pipe segment reports (i.e., pdf format), and the other information related to the pipes such as pipe age, corrosion, and the seismic zone is given in MS Excel from the Dept. of Engineering & Environmental Services, Shreveport, Louisiana Phase 3. There was no information related to loading, soil type, repair history in the documents, and these ratings were defined based on extensive information found in the literature (Data; Level; maps; Projections; Systems; Traffic). These Pipe Section reports contain different sections, as presented in Table 3. Each section contains text input by the inspector.

Table 3. Description of Pipe Segment Reports

| Section | Description |
| --- | --- |
| Physical characteristics | Information about the physical pipe properties (Ex: Diameter, Depth, Length) |
| Emergency Repair | Information about the Emergency Repair (Ex: Immediate Leakage Fixes) |
| Smoke Testing Assessment | Information about any smoke observed from pipes (Ex: Medium smoke observed emanating from cleanout) |
| CCTV Assessment | Information about the pipes using CCTV Camera (Ex: Multiple Defects) |
| Composite Assessment | Information about the Composite Material around the pipe |
| Criticality Assessment | Information about the risk value of the pipe (Ex: Medium) |
| Capacity | Information about the pipe Capacity |

We used Python programming to process the records of all the wastewater pipe documents to extract 12 specified variables from the pdf documents: Pipe ID, Pipe Diameter, Depth Category, Total Length (Feet), Existent Height (inches), Existent Material, Existent Lining Method, O&M score, Structural score, and Comprehensive Rating listed under Physical characteristics section into a .csv file. For our final data, we have combined the .csv file and MS Excel from the Dept. of Engineering & Environmental Services. The flow diagram of the data cleaning process to obtain the final data is shown in Figure 3. We then randomly selected 200 documents and manually checked the data to verify if the same data was extracted using Python programming. The extraction and retrieval of information by the program were compared to the results of the manual review. Sample Data extracted using the python programming into the .csv file is shown in Table 4.

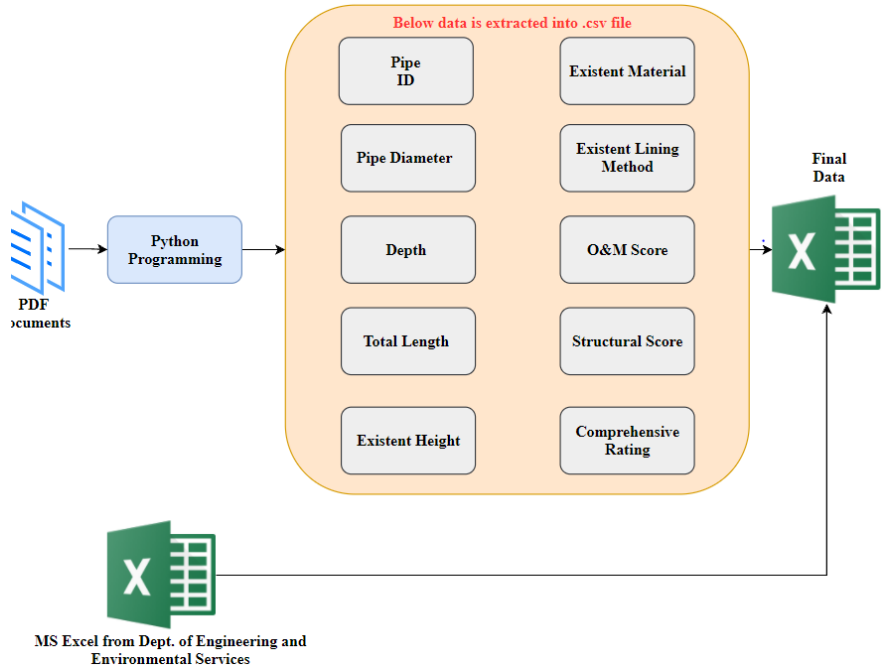

Figure 3: Final Data Cleaning Process.

Table 4. Pipe segment Repair data retrieved.

| Pipe ID | Pipe Diameter | Depth | Total Length (feet) | Existent Height (inches) | Existent Material | Existent Lining Method | O&M Score | Structural Score | Comprehensive Rating |
|---|---|---|---|---|---|---|---|---|---|
| 925 | 8 | 0-10 Feet | 86 | 10 | Vitrified Clay Pipe | None | 2 | 2 | 4 |
| 3336 | 8 | 10-15 Feet | 248 | 8 | Vitrified Clay Pipe | None | 4 | 4 | 2 |
| 1024 | 8 | 0-10 Feet | 414 | 10 | Vitrified Clay Pipe | None | 2 | 2 | 3 |

| 340 | 8 | 0-10 Feet | 154 | 8 | Vitrified Clay Pipe | None | 3 | 3 | 3 |
| 1454 | 8 | 10-15 Feet | 350 | 8 | Vitrified Clay Pipe | None | 4 | 4 | 4 |

**4.2 Data Preprocessing**

Data Preprocessing is when the data gets transformed, or encoded, such that the machine can quickly parse it. In this study, we included records with relevant data by removing 4.2% of records with inconsistent data, and 10% to 20% missing information info per pipe for further analysis. This step makes the training dataset cleaner and error free, which helps in improving the accuracy of the model. After all these analyses and verification of data, the final data collection included 3100 pipe segment reports, of which 1240 are considered for our analysis (Figure 4).

4.2.1. Missing values: It is very usual to have missing values in our dataset. It may have happened during data collection by the CCTV inspector. We eliminated 60 reports related to the few missing information such as pipe material, depth, or structural score.

4.2.2. Inconsistent values: We know that data can contain inconsistent values. For instance, the "Total Length" field contains the "Length Surveyed." It may be due to human error, or maybe the information was misread while being scanned from a handwritten form by the CCTV inspector. We have eliminated 70 reports related to the inconsistent values.

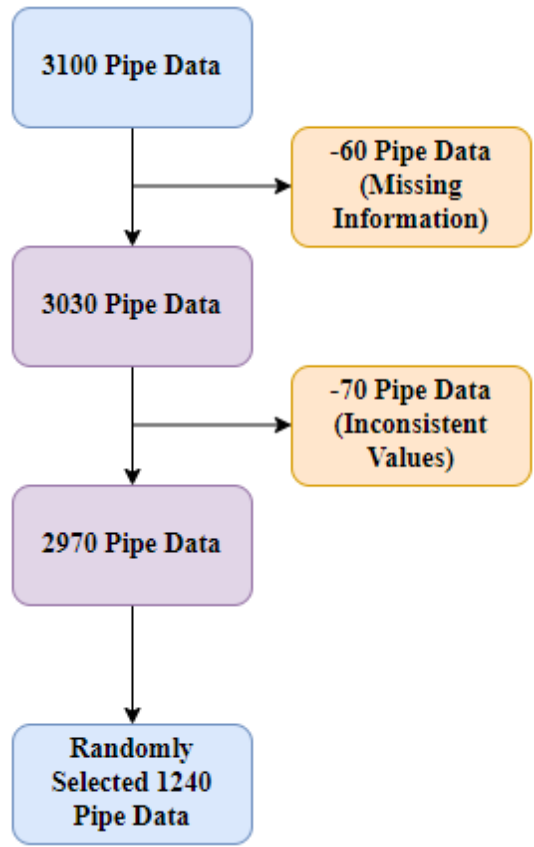

Figure 4: Process of Pipe Segment Reports

### 4.3 Comprehensive Rating Framework

The comprehensive rating model incorporates the well-established industry-standard condition rating method, the PACP developed by NASSCO (2001), from which the PACP structural and operation and maintenance scores are added to the comprehensive rating, as well as other physical characteristics, external characteristics, and hydraulic parameters to determine overall pipe condition score. The $K$-Nearest Neighbor model is used for this purpose. We didn't consider the geographical location of the pipe for our model implementation. A comprehensive rating framework is shown in Figure 5.

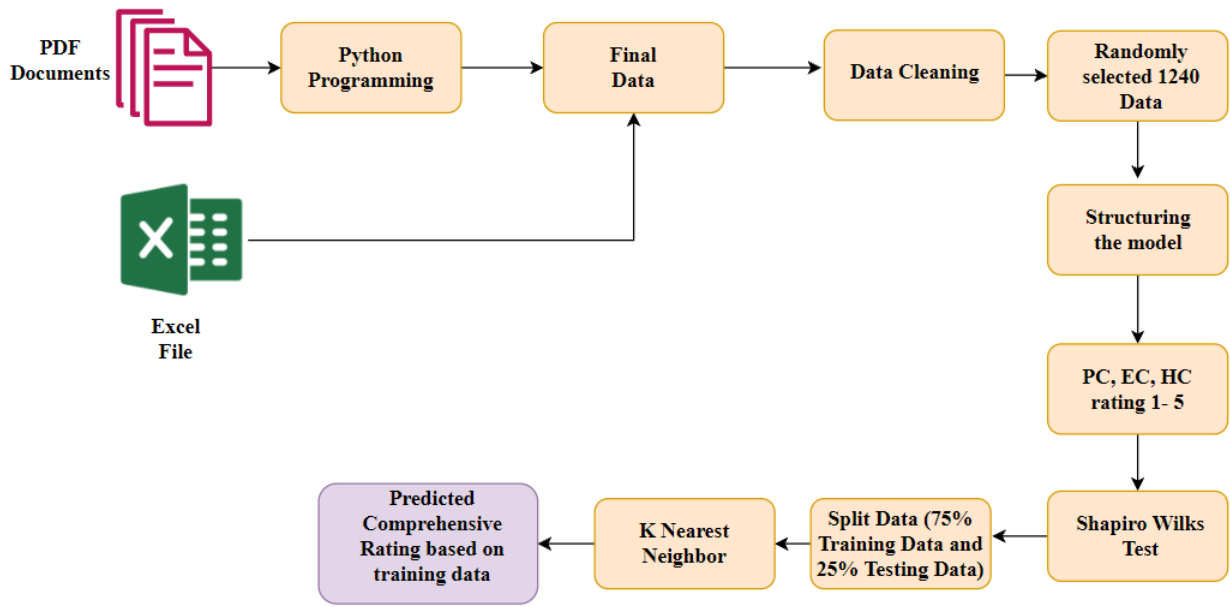

Figure 5. Comprehensive Rating framework

In the first step, the developed model is structured hierarchically. First, factors selected for this pipe deterioration process are selected from the literature and classified into three main characteristics: Physical characteristics (PC), External Characteristics (EC), and Hydraulic Characteristics (HC), as shown in Figure 6.

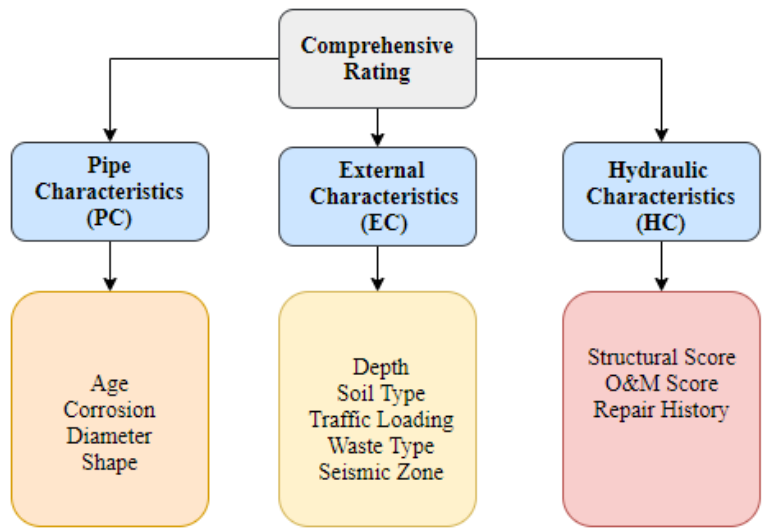

Figure 6: Structured model.

The factors summary is presented in Table 5. Factors' attributes, and the assigned ratings of physical characteristics (PC), external Characteristics (EC), and hydraulic Characteristics (HC) are presented in Table 6, Table 7, and Table 8.. All the PC, EC, HC ratings were defined based on extensive information found in the literature. The following factors are defined under the physical characteristics (PC) criteria: pipe age, material, diameter, length, and shape. Accordingly, as the pipe material ages, the degradation process becomes more significant. In addition, differing pipe materials are shown to have further deterioration and failure mechanisms; Several studies exist that have direct results related to the relationship between the pipe's diameter and its structural integrity(Balmer & Meers, 1981; Davies, Clarke, Whiter, & Cunningham, 2001; Ennaouri & Fuamba, 2013). Finally, different geometrical shapes will result in varying levels of deposits and degradation patterns.

Under the external characteristics (EC) criteria, the following factors are defined: burial depth, soil type, loading, waste carried, and seismic zone. The deep burial of the pipe results in increased soil overburden on the pipe. Next, the soil type refers to the surrounding soil that comes in direct contact with the pipe, which can impact the external pipe wall deterioration mechanism, mainly corrosive materials, hydrocarbons, etc., present in the soil. Traffic loads include all pedestrian and vehicle traffic above and in the proximity of the pipe, which impacts the overall integrity of the pipe. The type of waste carried can potentially corrode the internal pipe wall if highly corrosive. Including the seismic zone, factor ensures that any possible effects of seismic activities on the overall condition of the pipe are considered in the model.

Under the hydraulic characteristics (HC) criteria, the following factors are defined: PACP structural, PACP operations and maintenance (O&M) defects, and repair history. The PACP structural and O&M defect scores are on a scale of 1–5. The repair history gives information about the maintenance of pipes in the previous years.

Table 5. Comprehensive rating factors and description

| Criteria | Factor | Data Type | Description |
|---|---|---|---|
| Physical Characteristics (PC) | Pipe age (years) | Numeric | The time between pipe installation and inspection year and aged pipes have more issues. |
| | Pipe material | String | The pipe material includes various types of material, such as ceramic, glass, fiberglass, many metals, concrete, and plastic. |
| | Diameter(mm) | Numeric | Nominal pipe diameter and smaller diameters are not easy to access. |
| | Shape | String | Typically pipe shapes are circular but depending upon the project, shapes are changed. Circular shapes are easily accessed. |
| External Characteristics (EC) | Depth(feet) | Numeric | Higher-depth sewers are more challenging to access. |
| | Soil Type | String | Soil corrosiveness can impact the external pipe wall deterioration mechanism. |
| | Loading | String | A pipe failure on or near a high traffic area can significantly increase delays and detour distances that negatively affect the social impact. |
| | Waste Type | String | Waste materials carried in a pipe can impact the pipe failure by blocking, corrosion, etc. |
| | Seismic Zone | String | Zones with higher seismic activities can negatively impact the structure. |

| Hydraulic Characteristics (HC) | Structural Score | Numeric | The score is given based upon the structure alignment. |
|---|---|---|---|
| | O & M Score | Numeric | The score is given based upon the operational and maintenance. |
| | Repair History | String | Pipes with more maintenance can impact the final rating |

Table 6. Attributes factors rating for Physical Characteristics.

| Factor | Attribute | Ranking |
|---|---|---|
| Age Grade (years) | <10 | 1 |
| | ≥10 and <25 | 2 |
| | ≥25 and <40 | 3 |
| | ≥40 and <50 | 4 |
| | ≥50 years | 5 |
| Corrosion Resistance | Reinforced Plastic Pipe, Polyvinyl Chloride, Vitrified clay pipe, Polyethylene | 1 |
| | Cast Iron, Ductile Iron Pipe | 2 |
| | Reinforced Concrete Pipe, concrete pipe (non-reinforcement), Concrete Segments | 3 |
| | Not Known | 4 |
| | Other | 5 |
| Diameter | >=49 | 1 |
| | >31 and <=48 | 2 |
| | >18 and <=30 | 3 |
| | >11 and <= 18 | 4 |
| | <=11 | 5 |
| Shape | Circular | 1 |
| | Oval | 2 |
| | Horseshoe | 3 |
| | Semielliptical | 4 |

|  | Arch | 5 |

Table 7. Attributes factors rating for External Characteristics.

| Factor | Attribute | Ranking |
| --- | --- | --- |
| Depth | <= 10 Feet | 1 |
|  | > 10 and <= 15 Feet | 2 |
|  | > 15 and <= 20 Feet | 3 |
|  | > 20 and <= 25 Feet | 4 |
|  | > 25 Feet | 5 |
| Soil Type | Low corrosivity | 1 |
|  | Low to moderate corrosivity | 2 |
|  | Moderate corrosivity | 3 |
|  | Moderate-to-high corrosivity | 4 |
|  | High corrosivity | 5 |
| Loading | No traffic to very light traffic | 1 |
|  | Light traffic | 2 |
|  | Medium traffic | 3 |
|  | Moderate to heavy traffic | 4 |
|  | Heavy traffic | 5 |
| Waste Type | Mildly corrosive | 1 |
|  | Mildly to Moderate corrosive | 2 |
|  | Moderately corrosive | 3 |
|  | Moderately to highly corrosive | 4 |
|  | Highly corrosive | 5 |
| Seismic Zone | Zone 1 | 1 |
|  | Zone 2 | 2 |
|  | Zone 3 | 3 |
|  | Zone 4 | 4 |
|  | Zone 5 | 5 |

Table 8. Attributes factors rating for External Characteristics.

| Factor | Attribute | Ranking |
|---|---|---|
| Structural Score | 1 | 1 |
| | 2 | 2 |
| | 3 | 3 |
| | 4 | 4 |
| | 5 | 5 |
| O & M Score | 1 | 1 |
| | 2 | 2 |
| | 3 | 3 |
| | 4 | 4 |
| | 5 | 5 |
| Repair History | No maintenance | 1 |
| | Minor maintenance | 2 |
| | Moderate maintenance | 3 |
| | Significant maintenance | 4 |
| | Extreme maintenance | 5 |

## 4.4 Shapiro Wilks Test

In the next step, we determine the factors needed for our model. For this, we used the Shapiro Wilks test. Given an ordered random sample $z_1 < z_2 ... < z_n$, the original Shapiro-Wilk test (J. P. Royston, 1982; Royston, 1992) statistic is defined as

$$W = \frac{(\sum_{i=1}^{n} a_i z_{(i)})^2}{\sum_{i=1}^{n}(z_i - \bar{z})^2} \qquad (Eq.1)$$

where $z_{(i)}$ is the $i^{th}$ the smallest number in the sample, $\bar{z}$ is the sample mean and the coefficients $a_i$
are given by

$$(a_1, \ldots, a_n) = m^T V^{-1}/C,$$

where $C$ is a vector norm, i.e., $C = \|V^{-1}m\| = (m^T V^{-1} V^{-1} m)^{1/2}$ and vector $m$ is given by $m = (m_1, \ldots \ldots m_n)^T$ and $V$ is the covariance matrix.

$W$ is a number that ranges from 0 to 1. Small values of $W$ imply that normality is rejected, whereas a value of one show that the data is normal(J. Royston, 1982a, 1982b; J. P. Royston, 1982; Royston, 1992, 1995). A significance level of 0.05 is considered.

### 4.5 $K-$Nearest Neighbor

The next step is to build our model using $K-$ Nearest Neighbor ($K$-NN) classifier(Peterson, 2009). $K$-NN classifies the new data points based on the similarity measure of the earlier stored data points.

*Algorithm*:

Input: $E$: All factors, $K$: Chosen Number of Neighbors

*Output*: $C$: Mode of $K$ labels

*Begin*:

- Load the data.
- Initialize K to your chosen number of neighbors.
- For each testing data:
  - Calculate the distance between 25% of testing data $(x, y)$ with all 75% of the training data. (a, b) using Euclidean distance (ED) as shown in Equation 2.
  $$ED = \sqrt{(x - a)^2 + (y - b)^2)} \qquad (Eq.2)$$
  - Add the distance and the index of testing data to the ordered collection.
- Sort the ordered collection of distances and indices in ascending order by distances.
- Pick the first $K$ entries from the sorted collection.
- Get the labels of selected entries.
- Return the mode of $K$ labels.

End

## 5. Analysis

### 5.1 Shapiro Wilks Test

All variables consisting of the attributes are stored in a final spreadsheet. This final sheet is stored in JMP software(Jones & Sall, 2011; Kenett & Zacks, 2021) to perform the Shapiro Wilks test. The results for the Shapiro Wilks test are presented in Table 9. We hypothesized that all the factors are normally distributed; however, the distribution was not normal for diameter and seismic zone, so the hypothesis was rejected. The factors for which $w > \alpha = 0.05$ are considered in our modeling.

Table 9. Shapiro wilks test Result

| Factor | P value |
| --- | --- |
| Age | 0.980 |
| Corrosion | 0.883 |
| Diameter | 0.000 |
| Shape | 0.950 |
| Depth | 0.887 |
| Soil Type | 0.884 |
| Loading | 0.845 |
| Seismic Zone | 0.000 |
| Waste Type | 0.898 |
| Structural score | 0.535 |
| O&M score | 0.711 |
| Repair History | 0.712 |

### 5.2 *K*- Nearest Neighbor

We have divided the data into 75% training and 25% validation data, and the process is repeated several times with different values of K to reduce the errors and to make accurate predictions. We have finally chosen the value as $K = 7$. As the value of *K* is increased, our predictions become more stable and will have more accurate predictions up to a certain point. Figure 7 shows the graph of misclassification rate as a function of K for 25 and 30, and from both the graph, we see the lowest error is found at $K = 7$ with a value of 0.31290. We also checked for different values of *K*, such as 20 and 35, and we found the lowest value of misclassification rate at 7. So, we have used the value as $K = 7$ for better accuracy. Table 10 shows the count and misclassification rate for training data and testing data for different *K* values. Misclassification is slightly higher because of less training data for comprehensive rating 1 and slightly fewer training data for comprehensive rating 2 and 3 compared to comprehensive rating 4. This can be reduced when the model is trained with a more wide variety of data with different comprehensive ratings. In our scenario we didn't consider the entire dataset because we have more comprehensive ratings related to 3 and 4 than other comprehensive ratings.

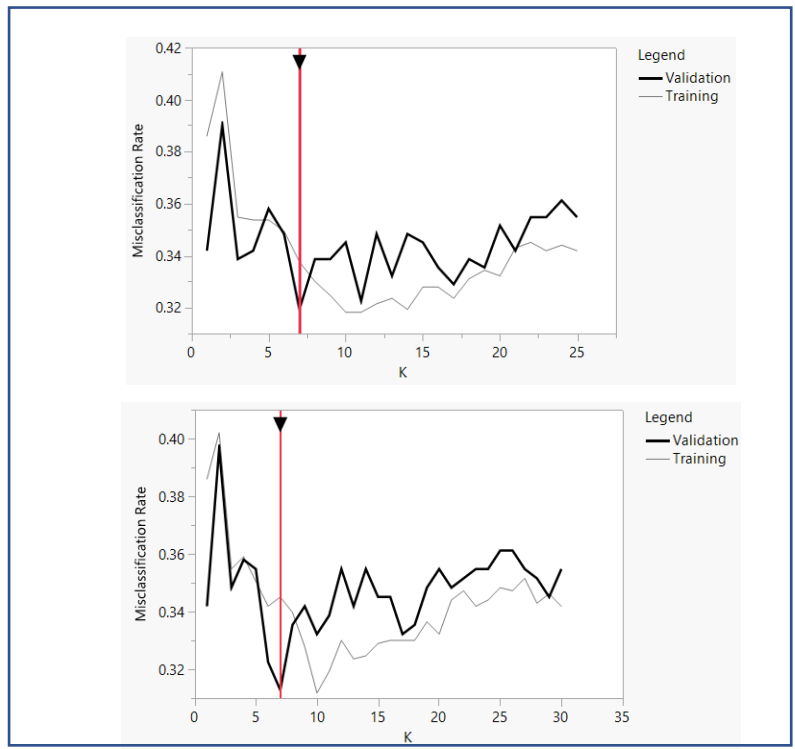

Figure 7. Misclassification rate as a function of *K*

Table 10. Misclassification rate for each *K*

|    | Training |                        | Validation |                        |
| --- | --- | --- | --- | --- |
| K  | Count    | Misclassification Rate | Count      | Misclassification Rate |
| 1  | 930      | 0.38602                | 310        | 0.34194                |
| 2  | 930      | 0.40215                | 310        | 0.39677                |
| 3  | 930      | 0.35484                | 310        | 0.34839                |
| 4  | 930      | 0.35914                | 310        | 0.35806                |
| 5  | 930      | 0.35054                | 310        | 0.35484                |
| 6  | 930      | 0.34194                | 310        | 0.32258                |
| 7  | 930      | 0.34516                | 310        | 0.31290                |
| 8  | 930      | 0.33978                | 310        | 0.33548                |
| 9  | 930      | 0.32796                | 310        | 0.34194                |
| 10 | 930      | 0.31183                | 310        | 0.33226                |
| 11 | 930      | 0.31935                | 310        | 0.33871                |
| 12 | 930      | 0.33011                | 310        | 0.35484                |
| 13 | 930      | 0.32366                | 310        | 0.34194                |
| 14 | 930      | 0.32473                | 310        | 0.35484                |
| 15 | 930      | 0.32903                | 310        | 0.34516                |
| 16 | 930      | 0.33011                | 310        | 0.34516                |
| 17 | 930      | 0.33011                | 310        | 0.33226                |
| 18 | 930      | 0.33011                | 310        | 0.33548                |
| 19 | 930      | 0.33656                | 310        | 0.34839                |
| 20 | 930      | 0.33226                | 310        | 0.35484                |
| 21 | 930      | 0.34409                | 310        | 0.34839                |
| 22 | 930      | 0.34731                | 310        | 0.35161                |
| 23 | 930      | 0.34194                | 310        | 0.35484                |
| 24 | 930      | 0.34409                | 310        | 0.35484                |
| 25 | 930      | 0.34839                | 310        | 0.36129                |
| 26 | 930      | 0.34731                | 310        | 0.36129                |
| 27 | 930      | 0.35161                | 310        | 0.35484                |
| 28 | 930      | 0.34301                | 310        | 0.35161                |

| 29 | 930 | 0.34624 | 310 | 0.34516 |
| 30 | 930 | 0.34194 | 310 | 0.35484 |

To proceed with *K*-NN calculation process, Euclidian distance is used to find the distance between each testing data to training data as shown in Equation 2. Table 11 shows the confusion matrix of validation data compared with the actual comprehensive ratings given by the inspector.

Table 11. Confusion Matrix for *K*-NN

| Predicted Comprehensive Rating | Actual Comprehensive Rating Count | | | | |
|---|---|---|---|---|---|
| | 1 | 2 | 3 | 4 | 5 |
| 1 | 22 | 10 | 0 | 0 | 0 |
| 2 | 3 | 46 | 5 | 6 | 0 |
| 3 | 0 | 7 | 70 | 12 | 5 |
| 4 | 0 | 0 | 9 | 51 | 6 |
| 5 | 0 | 2 | 8 | 10 | 38 |

We have compared our same data set using Analytical Hierarchy Process (AHP) and Naïve Bayes' Classifier (NBC) with the actual comprehensive ratings given by the inspector. Table 12 shows the confusion matrix of AHP, and Table 13 shows the confusion matrix of Naïve Bayes Classifier. The achieved overall accuracy of all the models is shown in Table 14.

Table 12. Confusion Matrix for AHP

| Predicted Comprehensive Rating | Actual Comprehensive Rating Count | | | | |
|---|---|---|---|---|---|
| | 1 | 2 | 3 | 4 | 5 |
| 1 | 3 | 12 | 21 | 17 | 4 |
| 2 | 8 | 1 | 17 | 27 | 12 |
| 3 | 4 | 16 | 18 | 22 | 21 |
| 4 | 6 | 24 | 15 | 6 | 11 |
| 5 | 4 | 12 | 21 | 7 | 1 |

Table 13. Confusion Matrix for Naïve Bayes Classifier

| Predicted | Actual Comprehensive Rating Count | | | | |
|---|---|---|---|---|---|

| Comprehensive Rating | 1 | 2 | 3 | 4 | 5 |
|---|---|---|---|---|---|
| 1 | 12 | 13 | 11 | 10 | 4 |
| 2 | 3 | 25 | 7 | 6 | 6 |
| 3 | 4 | 9 | 48 | 7 | 1 |
| 4 | 2 | 9 | 15 | 49 | 7 |
| 5 | 4 | 10 | 11 | 7 | 30 |

Table 14. Accuracy Comparison between KNN, Naïve Bayes Classifier, and AHP.

| Comprehensive Rating | KNN | AHP | Naïve Bayes Classifier |
|---|---|---|---|
| Accuracy | 73.23% | 9.35% | 52.90% |

Table 15 shows the accuracy, precision, recall, and F1 score for all 5 predicted comprehensive rating, and Equations 3 through 7 present the overall accuracy, precision, recall, and F1 score, respectively.

Table 15. Accuracy, Precision, Recall, and F1 score of all $K$ NN, AHP, NBC.

| Comprehensive Rating | Accuracy | | | Precision | | | Recall | | | F1 Score | | |
|---|---|---|---|---|---|---|---|---|---|---|---|---|
| | $K$ NN | AHP | NBC | $K$ NN | AHP | NBC | $K$ NN | AHP | NBC | $K$ NN | AHP | NBC |
| 1 | 95.81% | 75.48% | 83.44% | 0.69 | 0.05 | 0.24 | 0.88 | 0.12 | 0.48 | 0.77 | 0.07 | 0.32 |
| 2 | 89.35% | 58.71% | 79.55% | 0.77 | 0.01 | 0.53 | 0.71 | 0.01 | 0.38 | 0.74 | 0.01 | 0.44 |
| 3 | 85.16% | 55.81% | 78.90% | 0.74 | 0.22 | 0.70 | 076 | 0.20 | 0.52 | 0.75 | 0.21 | 0.60 |
| 4 | 86.13% | 58.39% | 79.55% | 0.77 | 0.09 | 0.59 | 0.65 | 0.07 | 0.61 | 0.70 | 0.08 | 0.60 |
| 5 | 90.00% | 70.32% | 83.77% | 0.66 | 0.02 | 0.48 | 0.78 | 0.02 | 0.63 | 0.71 | 0.02 | 0.55 |

$$Overall\ Accuracy = (\frac{correctly\ predicted}{total})*100\% \qquad (Eq.3)$$

$$Accuracy = (\frac{TP+TN}{TP+TN+FP+FN})*100\% \qquad (Eq.4)$$

$$Precision = \frac{TP}{TP+FP} \qquad (Eq.5)$$

$$Recall = \frac{TP}{TP+FN} \qquad (Eq.6)$$

$$F1\ Score = \frac{2TP}{2TP+FP+FN} \qquad (Eq.7)$$

where TP, FN, FP, and TN represent the number of true positives, false negatives, false positives, and true negatives, respectively.

In summary $K$-NN classifier with identified factors after Shapiro Wilks test is superior to Naïve Bayes classifier and AHP for classifying defect ratings based on the 10 factors and reducing the manual efforts of the inspector.

## 6. Conclusion

The proposed condition rating model assesses the overall state of degradation of the wastewater pipe, combining a series of physical characteristics, external characteristics, and hydraulic characteristics. The model considered 12 initial factors that contribute to the sewer pipe degradation and finally incorporated 10 factors based on the Shapiro Wilks test. Diameter and Seismic zone were the same for all the data; hence the data were not normally distributed, which resulted in the rejection of those two factors for our model A $K$-Nearest Neighbor ($K$-NN) model was used to find the pipe rating. To validate the model, the predicted Comprehensive ratings of our model were compared with actual comprehensive ratings, and our accuracy was 73.23% which is satisfactory. We also compared predicted comprehensive ratings using AHP and Naïve Bayes classifier with actual comprehensive ratings, and the accuracy were 9.35% and 52.90%, which shows the $K$-NN model is more accurate in predicting the comprehensive rating.

The main limitation of the study was the data. All the pipes' data had the same diameter and seismic zone. Therefore, more pipe from different geographic locations is needed to improve and convey more robustness to the obtained results.

In addition, for future research, more experimental applications to case studies are suggested for refining and improving the number of structural, operational, and hydraulic factors used in the model by considering more variety of data. By adding more factors, this method could be applied to any wastewater pipes to recognize the worst condition of wastewater pipes that need to be replaced immediately. In significantly less time by reducing many manual efforts. Secondly, integrating the consequence-of-failure model with comprehensive rating models can help utility managers prioritize critical investment needs in a more efficient manner

**References:**

Angkasuwansiri, T., & Sinha, S. (2015). Development of a robust wastewater pipe performance index. *Journal of Performance of Constructed Facilities*, *29*(1), 04014042.

Aprajita, F. (2018). *Guidelines for Implementing Risk-Based Asset Management Program to Effectively Manage Deterioration of Aging Drinking Water Pipelines, Valves and Hydrants* Virginia Tech].

Ariaratnam, S. T., El-Assaly, A., & Yang, Y. (2001). Assessment of infrastructure inspection needs using logistic models. *Journal of Infrastructure Systems*, *7*(4), 160-165.

Balmer, R., & Meers, K. (1981). Money Down the Drain 1. The Sewer Renewals Project of the Severn-Trent Water Authority. *Publ. Health Eng.*, *9*(1), 7-10.

Chughtai, F., & Zayed, T. (2008). Infrastructure condition prediction models for sustainable sewer pipelines. *Journal of Performance of Constructed Facilities*, *22*(5), 333-341.

Companies, N. A. o. S. S. *Pipeline Assessment and Certificattion Program (PACP): Reference manual, Volume 6.0.1,, Marriottsville, MD*

Data, U. W. *USGS water data for the nation*. Retrieved March 5, 2019 from https://waterdata.usgs.gov/nwis

Davies, J., Clarke, B., Whiter, J., & Cunningham, R. (2001). Factors influencing the structural deterioration and collapse of rigid sewer pipes. *Urban water*, *3*(1-2), 73-89.

Davies, J., Clarke, B., Whiter, J., Cunningham, R., & Leidi, A. (2001). The structural condition of rigid sewer pipes: a statistical investigation. *Urban water*, *3*(4), 277-286.

Ennaouri, I., & Fuamba, M. (2013). New integrated condition-assessment model for combined storm-sewer systems. *Journal of Water Resources Planning and Management*, *139*(1), 53-64.

Jones, B., & Sall, J. (2011). JMP statistical discovery software. *Wiley Interdisciplinary Reviews: Computational Statistics*, *3*(3), 188-194.

Kenett, R. S., & Zacks, S. (2021). *Modern industrial statistics: With applications in R, MINITAB, and JMP*. John Wiley & Sons.

Khazraeializadeh, S., Gay, L. F., & Bayat, A. (2014). Comparative analysis of sewer physical condition grading protocols for the City of Edmonton. *Canadian Journal of Civil Engineering*, *41*(9), 811-818.

Lester, J., & Farrar, D. (1979). *An Examination of the Defects Observed in Six Kilometres of Sewers*.

Level, U. G. W. *Groundwater levels for the nation*. Retrieved March 5, 2019 from https://nwis.waterdata.usgs.gov/usa/nwis/gwlevels.

maps, U. S. h. *Seismic hazard maps and site-specific data.* Retrieved April 7,2018 from https://earthquake.usgs.gov/hazards/hazmaps/

McDonald, S. E., & Zhao, J. Q. (2020). Condition assessment and rehabilitation of large sewers. In *Underground Infrastructure Research* (pp. 361-369). CRC Press.

Najafi, M., & Kulandaivel, G. (2005). Pipeline condition prediction using neural network models. In *Pipelines 2005: Optimizing Pipeline Design, Operations, and Maintenance in Today's Economy* (pp. 767-781).

O'REILLY, M., Rosbrook, R., Cox, G., & McCloskey, A. (1989). *Analysis of defects in 180km of pipe sewers in Southern Water Authority* (0266-5247).

Opila, M. C. (2011). *Structural condition scoring of buried sewer pipes for risk-based decision making*. University of Delaware.

Opila, M. C., & Attoh-Okine, N. (2011). Novel approach in pipe condition scoring. *Journal of Pipeline Systems Engineering and Practice*, *2*(3), 82-90.

PACP. (2021). *PACP Condition Grades*. Retrieved October 12 from https://nassco.org/resource/pacp-condition-grades-and-their-proper-application/

Peterson, L. E. (2009). K-nearest neighbor. *Scholarpedia*, *4*(2), 1883.

Projections, U. L. C. *United States land cover projections*. Retrieved March 5, 2019 from https://www.sciencebase.gov/catalog/item/5b96c2f9e4b0702d0e826f6d

Rehab, P. S. C. Retrieved October 12 from http://www.ohiowea.org/docs/03_Sewer_Conditions_Rehab.pdf

Royston, J. (1982a). Algorithm AS 177: Expected normal order statistics (exact and approximate). *Journal of the royal statistical society. Series C (Applied statistics)*, *31*(2), 161-165.

Royston, J. (1982b). Algorithm AS 181: the W test for normality. *Applied Statistics*, 176-180.

Royston, J. P. (1982). An extension of Shapiro and Wilk's W test for normality to large samples. *Journal of the Royal Statistical Society: Series C (Applied Statistics)*, *31*(2), 115-124.

Royston, P. (1992). Approximating the Shapiro-Wilk W-test for non-normality. *Statistics and computing*, *2*(3), 117-119.

Royston, P. (1995). Remark AS R94: A remark on algorithm AS 181: The W-test for normality. *Journal of the royal statistical society. Series C (Applied statistics)*, *44*(4), 547-551.

Sai Nethra Betgeri, J. C. M., David B. Smith. (2021). Comparison of Sewer Conditions Ratings with Repair Recommendation Reports. *North American Society for Trenchless Technology (NASTT) 2021*,

Systems, U. W. *Small wastewater systems research*. Retrieved November 26, 2018 from https://www.epa.gov/water-research/small-wastewater-systems-research-0.\

Traffic, U. *US traffic volume data*. Retrieved March 3, 2019 from https://www.transportation.gov/

Vadyala, S. R., & Betgeri, S. N. (2021). Physics-Informed Neural Network Method for Solving One-Dimensional Advection Equation Using PyTorch. *arXiv preprint arXiv:2103.09662*.

Vadyala, S. R., Betgeri, S. N., Sherer, E. A., & Amritphale, A. (2021). Prediction of the number of covid-19 confirmed cases based on k-means-lstm. *Array*, *11*, 100085.

Vladeanu, G., & Matthews, J. (2019). Wastewater pipe condition rating model using multicriteria decision analysis. *Journal of Water Resources Planning and Management*, *145*(12), 04019058.

Vladeanu, G. J., & Matthews, J. C. (2019). Consequence-of-failure model for risk-based asset management of wastewater pipes using AHP. *Journal of Pipeline Systems Engineering and Practice*, *10*(2), 04019005.

Wirahadikusumah, R., Abraham, D., & Iseley, T. (2001). Challenging issues in modeling deterioration of combined sewers. *Journal of Infrastructure Systems*, *7*(2), 77-84.

Yan, J., & Vairavamoorthy, K. (2003). Fuzzy approach for pipe condition assessment. In *New pipeline technologies, security, and safety* (pp. 466-476).